\documentclass{article}
\usepackage{arxiv}

\usepackage{graphicx}
\usepackage{amstext}
\usepackage{amssymb}
\usepackage{amsmath}
\usepackage{pgfplots}
\usepackage{amsfonts}
\usepackage{hyperref}
\usepackage{url}
\usepackage{makecell}
\usepackage{booktabs}
\usepackage{multirow}
\usepackage{subcaption}
\usepackage{float}
\usepackage{tablefootnote}
\usepackage{color}
\usepackage{numprint}
\usepackage{nicefrac}       
\usepackage{microtype}      
\usepackage{natbib}
\usetikzlibrary{patterns}
\usetikzlibrary{decorations.pathreplacing}
\usetikzlibrary{pgfplots.groupplots}
\pgfplotsset{compat=1.3}

\def\x{{\mathbf x}}
\def\y{{\mathbf y}}
\def\yhat{\hat \y}

\def\devclean{\texttt{dev-clean}}
\def\devother{\texttt{dev-other}}
\def\testclean{\texttt{test-clean}}
\def\testother{\texttt{test-other}}

\usepackage[colorinlistoftodos, textwidth=20mm]{todonotes}
\definecolor{citrine}{rgb}{0.89, 0.82, 0.04}
\definecolor{blued}{RGB}{70,197,221}
\definecolor{plum}{RGB}{221,160,221}

\newcommand{\librivox}{\textsc{LibriVox}}
\newcommand{\librilight}{\textsc{LibriLight}}
\newcommand{\librispeech}{\textsc{LibriSpeech}}

\title{End-to-End ASR: from Supervised to Semi-Supervised Learning with Modern Architectures}

\author{%
Gabriel Synnaeve\thanks{Equal contribution.} \\
Facebook, NYC \\
\texttt{gab@fb.com} \\
\And
Qiantong Xu$^*$ \\
Facebook, Menlo Park \\
\texttt{qiantong@fb.com} \\
\And 
Jacob Kahn$^*$ \\
Facebook, Menlo Park \\
\texttt{jacobkahn@fb.com} \\
\And 
Tatiana Likhomanenko$^*$ \\
Facebook, Menlo Park \\
\texttt{antares@fb.com} \\
\And
Edouard Grave$^*$ \\
Facebook, Paris \\
\texttt{egrave@fb.com} \\
\And
Vineel Pratap \\
Facebook, Menlo Park \\
\texttt{vineelkpratap@fb.com} \\
\And
Anuroop Sriram \\
Facebook, Menlo Park \\
\texttt{anuroops@fb.com} \\
\And
Vitaliy Liptchinsky \\
Facebook, Menlo Park \\
\texttt{vitaliy888@fb.com} \\
\And
Ronan Collobert$^*$ \\
Facebook, Menlo Park \\
\texttt{locronan@fb.com} \\
}
\begin{document}

\maketitle
\setcounter{footnote}{0}

\begin{abstract}
We study pseudo-labeling for the semi-supervised training of ResNet, Time-Depth Separable ConvNets, and Transformers for speech recognition, with either CTC or Seq2Seq loss functions.
We perform experiments on the standard \librispeech~dataset, and leverage additional unlabeled data from \librivox~through pseudo-labeling.
We show that while Transformer-based acoustic models have superior performance with the supervised dataset alone, semi-supervision improves all models across architectures and loss functions and bridges much of the performance gaps between them.
In doing so, we reach a new state-of-the-art for end-to-end acoustic models decoded with an external language model in the standard supervised learning setting, and a new absolute state-of-the-art with semi-supervised training.
Finally, we study the effect of leveraging different amounts of unlabeled audio, propose several ways of evaluating the characteristics of unlabeled audio which improve acoustic modeling, and show that acoustic models trained with more audio rely less on external language models.
\end{abstract}

\section{Introduction}
\label{introduction}
\label{sec:intro}
End-to-end speech recognition models are simpler to implement and train than bootstrapped systems. Even given recent promising results from these systems, best-results for common benchmarks are still dominated by classical ASR models; systems requiring force alignment may leave some performance aside for each training step. We set out to study end-to-end systems on \librispeech~\cite{panayotov2015librispeech} and, without any algorithmic contribution, see if they can be made to perform as well as more complex training pipelines. The difficulties involved in properly optimizing acoustic models with Connectionist Temporal Classification (CTC) \cite{graves2006connectionist} or sequence-to-sequence (Seq2Seq) \cite{sutskever2014sequence} (v.s. cross-entropy, for instance) combined with more readily-available regularization techniques for classical pipelines make this comparison challenging. Our best acoustic models nonetheless reach 5.17\% WER on \testother, showing that end-to-end models can compete with traditional pipelines.

As in other domains, self and semi-supervised learning in ASR, where a pretrained network generates and trains on its own labels, yields improvements \cite{vesely2017semi}. In end-to-end ASR, pseudo-labeling and self-training can be quite effective, and its effectiveness is further improved when more data is available \cite{kahn2019self}. In this setting, we train a model on \librispeech, then use that model in conjunction with a language model to generate pseudo-labels from unlabeled audio. We show that with this training scheme, our results \textit{without an external language model} (LM) reach state-of-the-art results that \textit{use an external language model}, with 2.28\% and 4.88\% Word Error Rate (WER) on \testclean~and \testother~ respectively. With LM beam-search decoding and rescoring, we reach 2.09\% and 4.11\% WER on the test set.

\begin{figure}[!t]
\centering
\caption{\label{fig:resultsgraphdev}WERs on \devother~across AM architectures and loss functions. \textit{Left}: WERs of different models trained on \librispeech~with and without beam-search decoding ("no LM" refers to the greedy decoding). Transformer AM architectures outperform others by a large margin. \textit{Right}: WERs of models trained on \librivox. All models trained on \librivox~significantly outperform their \librispeech~counterparts. The gap between Transformer AMs and other models is much smaller with \librivox~data.}
\begin{tikzpicture}
    \begin{groupplot}[group style = {group size = 2 by 1, horizontal sep = 15pt}, width =4.7cm, height = 6.0cm]
        \nextgroupplot[ 
 	        xticklabels={ResNet, TDS, Tran.},
         	xtick={1,2,3},
            x tick label style={ yshift=-.5ex, xshift=-0.ex,font=\small},
	        ylabel=Word Error Rate,
	        ymajorgrids,
            y tick label style={ font=\small},
            y label style={ font=\small},
	        enlarge x limits=0.2,
	        ymin=3,ymax=12,
            legend style={at={(2.0,-0.15)},legend columns=2,font=\small}
            ]
            \addplot[red, draw=none, mark=*] coordinates
            {(1, 10.13) (2, 11.16) (3, 7.31) };
            \addplot[red, draw=none, mark=triangle*] coordinates
            {(1, 9.89) (2, 8.19) (3, 6.67) };
            
            \addplot[brown, draw=none, mark=*] coordinates
            {(1, 8.56) (2, 9.18) (3, 6.20) };
            \addplot[brown, draw=none, mark=triangle*] coordinates
            {(1, 8.69) (2, 7.01) (3, 5.81) };
            
            \addplot[blue, draw=none, mark=*] coordinates
            {(1, 7.50) (2, 7.52) (3, 5.29) };
            \addplot[blue, draw=none, mark=triangle*] coordinates
            {(1, 7.86) (2, 6.30) (3, 5.20) };
            \legend{CTC + no LM, S2S + no LM, CTC + ngram, S2S + ngram, CTC + GCNN, S2S + GCNN}
        \nextgroupplot[ 
 	        xticklabels={ResNet, TDS, Tran.},
         	xtick={1,2,3},
         	ymajorgrids,
            x tick label style={yshift=-.5ex, xshift=-0.ex,font=\small},
	        enlarge x limits=0.2,
	        ymin=3,ymax=12,
            ]
            \addplot[red, draw=none, mark=*] coordinates
            {(1, 5.54) (2, 5.59) (3, 5.00) };
            \addplot[red, draw=none, mark=triangle*] coordinates
            {(1, 5.29) (2, 4.78) (3, 4.59) };
            
            \addplot[brown, draw=none, mark=square*] coordinates
            {(1, 5.45) (2, 5.86) (3, 4.94) };
            \addplot[brown, draw=none, mark=triangle*] coordinates
            {(1, 5.28) (2, 4.80) (3, 4.53) };
            
            \addplot[blue, draw=none, mark=*] coordinates
            {(1, 4.64) (2, 4.82) (3, 4.27) };
            \addplot[blue, draw=none, mark=triangle*] coordinates
            {(1, 4.91) (2, 4.21) (3, 3.95) };
    \end{groupplot}
\end{tikzpicture}
\end{figure}
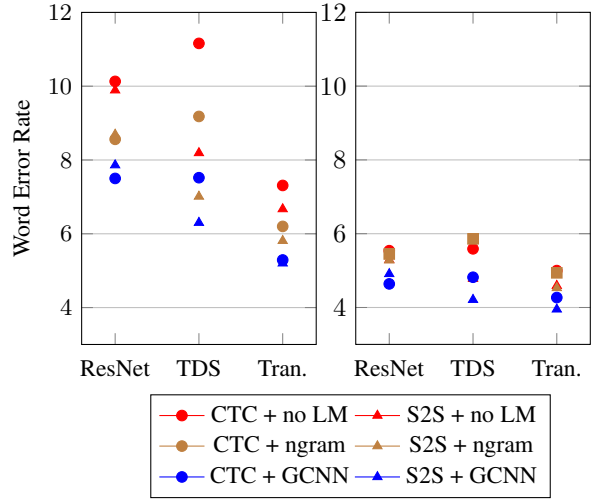

While many advances in end-to-end ASR come as the result of neural architecture search \cite{prabhavalkar2017comparison,zhou2018comparison,chiu2018state}, we additionally show that simple semi-supervision via pseudo-labeling significantly bridges the performance gap between a variety of different model architectures and loss functions, as shown in Figure~\ref{fig:resultsgraphdev}. In particular, with enough unlabeled audio, Transformer, ResNet, and depthwise-separable convolution-based acoustic models give similar performance with both CTC and Seq2Seq loss functions, suggesting that new techniques in semi-supervision may facilitate equally-significant gains in ASR performance while being applicable to a multitude of end-to-end setups.

\section{Models}
\label{sec:models}
\subsection{Acoustic Models}
\label{sec:acousticmodels}
In this section, we present the three families of acoustic models (AMs) studied. All AMs output probability distributions over tokens. In particular, we use a set of 10k word pieces \cite{schuster2012wpm,kudo2018sentencepiece} generated from the \textit{SentencePiece} toolkit\footnote{\url{https://github.com/google/sentencepiece}}. The choice to use a fixed set of 10k word pieces is made for the simplicity of the comparative study, not the result of a limitation. Similarly, all AMs take 80-channel log-mel filterbanks as input, with STFTs computed on Hamming windows strided by 10ms. This window size is 25ms for Transformer models and 30ms for TDS and ResNet models. All models are trained end-to-end with either CTC or Seq2Seq loss. Given the huge difference between the amounts of data, we prepare two sets of architectures: one for training only on labeled \librispeech~and one for unlabeled \librivox. 

\textbf{ResNet Acoustic Models.}~
\label{sec:resnet}
ResNets were first introduced in the domain of computer vision~\cite{he2016resnets} and have since been successfully applied to speech recognition~\cite{xion2017,saon2017,li2019jasper,wang2017residual}. ResNets are composed of several blocks of convolutions (in our case only 1-D convolutions), with skip connections to reduce the effect of vanishing gradients in deep neural networks. In particular, our ResNet encoder includes 42 convolutional layers each with a kernel size of $3$. The encoder first maps the input to an embedding space of size $1024$ using a single convolutional layer with stride $2$; $12$ blocks of three 1-D convolutions each follow. Each of the convolutional layers is followed by ReLU, dropout and LayerNorm~\cite{ba2016layer}. Both the dropout and the number of hidden units increases with the depth of the network. Specific convolution layers are inserted between ResNet blocks in order to upsample when the hidden representation size increases. Our architecture performs significant pooling with respect to the input ($16$ frames in total, equating to $160$ milliseconds) -- in addition to the first strided convolutional layer, $3$ max pooling layers (each with stride $2$) are distributed across the depth of the network (after blocks $3$, $7$ and $10$). Nearly identical encoder architectures are used in front of CTC and Seq2Seq loss functions; the Seq2Seq encoder has its last bottleneck layer removed and lower dropout in deeper layers. The Seq2Seq self-attention decoder for the ResNet architecture is the same as that used with the TDS convolutional AM described below. To better fit the unlabeled data, we increase the model size by increasing the number of channels in each convolution layer. 

\textbf{Time-Depth Separable (TDS) Convolution Acoustic Models.}~
\label{sec:TDS}
We extend the TDS block \cite{hannun2019TDS} (which is composed of one 2-D convolution layer and two fully-connected layers with ReLU, LayerNorm and residual connections in between), by increasing the number of channels in the feature maps spanning the two internal fully-connected layers by a factor $F > 1$, so as to increase model capacity. Following \cite{hannun2019TDS}, $3$ sub-sampling layers, i.e. 1-D convolution layers with stride $2$, are adopted to ensure an optimal context size for the encoder. 
For training with only labeled data, we have three groups of TDS blocks with $F = 3$ after each sub-sampling layers. There are $5$, $6$, and $10$ blocks in each group, containing $10$, $14$, and $18$ channels, respectively. To increase model capacity for unlabeled data, the three groups of TDS blocks, having fewer $4$, $5$, and $6$ blocks and $F = 2$ in each, are equipped with much larger $16$, $32$, and $48$ channels. 
All convolutions in both TDS and sub-sampling layers have kernel shapes of $21 \times 1$. Identical encoder architectures are shared between CTC and Seq2Seq. 

Our Seq2Seq self-attention decoder performs $R$ rounds of attention through the same $N$-layers of RNN-GRU each with a hidden unit size of $512$ in conjunction with the same efficient key-value attention as in \cite{hannun2019TDS,vaswani2017attention}: 
\begin{equation}
\mathbf{S}^r_t = \textsc{SoftMax}(\frac{1}{\sqrt{d}} \mathbf{K}^\top \mathbf{Q}_t^{r-1})\mathbf{V},
\label{eq:selfattn}
\end{equation}
where $[\mathbf{K}, \mathbf{V}]$ is $512$-dimensional encoder activation and $\mathbf{Q}_t^r = g(\mathbf{Q}_{t-1}^{r}, \mathbf{Q}_{t}^{r-1}) + \mathbf{S}^{r}_t$ is the query vector at time $t$ in round $r$, generated by the GRU $g(\cdot)$. The initial $\mathbf{Q}^0_t$ is a 512-dimensional token embedding, and the final $\mathbf{Q}^R_t$ is linearly projected to output classes for token classification. 
In our experiments, $N$ and $R$ are both set to either $2$ or $3$ based on validation performance. We use dropout in all TDS blocks and GRUs to prevent overfitting.

\textbf{Transformer-Based Acoustic Models.}~
\label{sec:transformers}
Our transformer-based acoustic models have a small front-end: $3$~(\librispeech~AMs) or $6$~(\librivox~AM) layers of 1-D convolutions each of kernel width $3$~and respective input and output sizes $(80,D_c)$, $(D_c/2,D_c)$, [$(D_c/2,D_c)$, $(D_c/2,D_c)$, $(D_c/2,D_c)$,] $(D_c/2,D_{tr}\times2)$, with $D_c=1024$ or $2048$. Each convolution is followed by a GLU activation function \cite{dauphin2017gcnn} and are striding by $2$ each (for $3$ consecutive layers), or every other layer (for $6$ layers). The output of the front-end for all models is thus strided by $8$ frames (80 ms). After the front-end, each Transformer block has $4$ attention heads followed by a feedforward network (FFN) with one hidden layer and a ReLU non-linearity. There are two configurations of Transformer blocks: one $24$ layer configuration (only for the \librispeech~CTC AM) with dimension $D_{tr}=1024$ for the self-attention and $4096$ for the FFN, and one $36$ layer configuration with dimension $D_{tr}=768$ for the self-attention and $3072$ for the FFN. Specifically, given a sequence of $T$ vectors of dimension $d$, the input is represented by the matrix $\mathbf{H^0} \in \mathbb{R}^{d \times T}$, following exactly \cite{vaswani2017attention}:
\begin{align*}
\mathbf{Z}^{i} & = \textsc{Norm}(\textsc{SelfAttention}(\mathbf{H}^{i-1}) + \mathbf{H}^{i-1}), \\
\mathbf{H}^{i} & = \textsc{Norm}(\textsc{FFN}(\mathbf{Z}^i) + \mathbf{Z}^i),
\end{align*}
where $\mathbf{Z}$ is the output of the self-attention layer, with a skip connection, and $\mathbf{H}$ is the output of the FFN layer, with a skip connection. As is standard: our $\textsc{Norm}$ is LayerNorm, and self-attention is defined as in Eq.~\ref{eq:selfattn}, but with $\mathbf{K}=\mathbf{W}_K\mathbf{H}$, $\mathbf{Q}=\mathbf{W}_Q\mathbf{H}$, and $\mathbf{V}=\mathbf{W}_V\mathbf{H}$.
For CTC-trained models, the output of the encoder $\mathbf{H}^{L_e}$ is followed by a linear layer to the output classes.
For Seq2Seq models, we have an additional decoder, which is a stack of 6 Transformers with encoding dimension 256 and 4 attention heads. The probability distribution of the transcription is factorized as:
\begin{equation}
p(y_1, ..., y_n) = \prod_{i=1}^n p(y_i \ | \ y_0, ..., y_{i-1}, \mathbf{H}^{L_e}),
\end{equation}
where $y_0$ is a special symbol indicating the beginning of the transcription. For all layers (encoder and decoder -- when present), we use dropout on the self-attention. We also use layer drop \cite{fan2019reducing}, dropping entire layers at the FFN level.

\subsection{Language Models}
\label{sec:languagemodels}

In this section, we present external language models (LMs) used in beam-search decoding. We consider $n$-gram LMs as well as convolutional~\cite{dauphin2017gcnn} (GCNN) and Transformer-based LMs. For $n$-gram and GCNN LMs, we train both word-based and word-piece models, and only a word-level Transformer LM. All word-piece LMs are trained on the set of 10k word pieces as outlined in Section~\ref{sec:acousticmodels}. This ensures that the set of word pieces is consistent across both of the output distributions of the acoustic models and the candidates the language model scores during beam-search decoding.

For the word-piece and word-level GCNN models, we use the GCNN-14B architecture from~\cite{dauphin2017gcnn} with embedding size $1024$ and dropout 0.1. The word-level Transformer LM has the same architecture as~\cite{baevski2018adaptive}'s \textit{Google Billion Words} model; we use 16 attention heads and 20 decoder layers with embedding, input and output dimensions of $1280$ and $6144$ for the FFN with dropout of 0.1.

\section{Dataset and Language Models for Pseudo-Labeling}
\label{sec:librivoxdataset}

\subsection{Unlabeled Audio Dataset Preparation}
\label{sec:librivoxdatasetprep}

\librivox\footnote{\url{https://librivox.org}}~is a large collection of freely-available audio books. Using tools provided with the \librilight~dataset~\cite{librilight}, we select 72K hours of read speech from English book listings and run several preprocessing steps. After filtering samples to remove readings of duplicate text and corrupted audio, we remove all audio for which the speaker has overlap with a sample in \librispeech. We run voice activity detection (VAD) using the wav2letter++ framework~\cite{pratap2018wav2letter} on the resulting collection of audio with a CTC model trained on \librispeech, and segment the result into chunks no greater than 36 seconds; the resulting audio corpus contains 53.8K hours of read speech.

We then generate pseudo-labels for this audio using the recipe described in~\cite{kahn2019self}. To generate the pseudo-labels, we use a Transformer AM trained on \librispeech~with CTC loss that achieves a 6.20\% WER on dev-other when decoded with a 4-gram word LM -- the same model as is listed in Table~\ref{tab:libriWERAppendix} in Appendix. We pseudo-label all audio using this AM and run beam-search decoding with a 4-gram word LM described below in Section \ref{sec:librivoxlmcorpus}.

\subsection{Text Corpus Preparation and $n$-gram LM Training}
\label{sec:librivoxlmcorpus}

The \librispeech~language model corpus\footnote{\url{https://www.openslr.org/11/}} contains text from $14500$ public domain books taken from the Gutenberg project\footnote{\url{https://www.gutenberg.org/}}. Given that pseudo-labels are generated with a beam-search decoding procedure that integrates a language model, it is important that the corpus used to train the language model does not have overlap with the unlabeled audio, else information about the ground truth labels for that unlabeled audio may be explicitly embedded in the LM. We remove all text from the \librispeech~language model training corpus that is ground truth for any of the unlabeled audio from the subset of \librivox.

To do so, we follow several steps. Firstly, we filter out all books from the \librispeech~LM corpus with IDs present in \librivox. Secondly, after normalizing all titles (removing punctuation, casing, and non-alphanumeric tokens), we remove all titles with zero Levenshtein distance between titles from the \librivox~and the \librispeech~LM corpuses. We use a Levenshtein metric over words rather than tokens for improved performance. We then find titles with nonzero but low similarity scores isolated via the following conditions. Given two book title strings $s_1$ and $s_2$, and constants $\alpha$ and~$\beta$:
\[
\max\{|s_1|, |s_2|\} - \min\{|s_1|, |s_2|\} < \alpha \cdot \min\{|s_1|, |s_2|\} \And
\]
\[
\text{Levenshtein}(s_1, s_2) \leq \beta \cdot \max\{|s_1|, |s_2|\}
\]
where notation $|s|$ refers to the number of words in the string $|s|$, and $0.75$ and $0.3$ were used as values for $\alpha$ and $\beta$, respectively. These constants are found empirically to remove obviously different titles and to have reasonable number of pairs (~10k) for further manual check. Titles that are manually matched are removed to create the final corpus; 13\% of the original \librispeech-LM corpus was filtered with the aforementioned steps.

Before training LMs, we normalize the filtered corpus so as to mimic the original normalization procedure found in \librispeech. 88\% of our normalized/filtered corpus has identical normalized text compared to the original \librispeech~LM corpus. As a result of our using a different tokenizer, sentence boundaries may differ across corpuses, as may abbreviations (e.g. we map `\&c' to `et cetera').

A $4$-gram language model is trained with the resulting corpus using the KenLM toolkit \cite{heafield2011kenlm} and the top 200k words as vocabulary. The model is trained without pruning (183k of the top 200k words are the same as the original \librispeech~LM corpus). This model is then used at beam-search decoding time in conjunction with an acoustic model trained on \librispeech~to generate pseudo-labels on the subset of \librivox~detailed in Section~\ref{sec:librivoxdataset}. During beam-search decoding we use a lexicon which is constructed from the \librispeech~ train sets only.

The perplexity difference between the 4-gram LM trained on the filtered corpus and the 4-gram LM trained on original \librispeech~LM corpus is small. The word perplexity of each model is shown in Table~\ref{tab:libriPPL}. Beam-search decoding of a Transformer AM trained on \librispeech~with an LM trained on the filtered corpus results in only a 0.05\% absolute WER regression on \devother~compared to decoding with an $n$-gram trained on the full corpus.

\section{Decoding}

Decoding is designed to select the best transcription by leveraging both the posteriors of an acoustic model (AM) and the perplexity of a language model (LM). We perform one-pass beam-search decoding with a single external LM. Optionally, to further improve performance, we use stronger NN-based LMs to rescore the beam.

\subsection{Beam-search Decoder}
In our experiments, we use lexicon-based and lexicon-free beam-search decoders following \cite{collobert2016wav2letter,likhomanenko2019needs} with either $n$-gram or GCNN LMs. The lexicon-based decoder, whose search space is limited to the words in the lexicon, is used for CTC models with a word-level LM. The lexicon-free decoder is capable of generating words with arbitrary spelling and is used for S2S models with a word-piece LM. The decoder takes as input posteriors from an acoustic model, a prefix trie built on a lexicon, and an external LM. We tune the language model weight $\alpha$ and the word insertion penalty $\beta$ on validation sets (\devclean~ and \devother). The decoder outputs a transcription $\yhat$ that maximizes: 
\begin{equation*}
\log P_{AM}(\yhat | \x) + \alpha \log P_{LM}(\yhat) + \beta|\yhat|.
\end{equation*}
To stabilize the Seq2Seq beam search, we introduce an EOS-penalty $\gamma$ to hypothesis that have finished in an end-of-sentence token. $\gamma$ is tuned together with other hyper-parameters and our experiments show that this strategy effectively prevents the decoder from early-stopping. To improve decoding efficiency, we also incorporate the thresholding technique in \cite{hannun2019TDS} and strategies mentioned in \cite{zeghidour2018fullyconv} including hypothesis merging, score caching, and batched LM forwarding. For CTC decoding, following \cite{park2018fully}, only the blank token is considered if its posterior probability is greater than 0.95.

\subsection{Rescoring}\label{sec:rescoring}
After acquiring the transcriptions of the $N$-best hypotheses from the one-pass beam-search decoder, we use an external word-level GCNN LM and a Transformer LM to evaluate their log-probabilities, denoted as $\log P_1(\yhat)$ and $\log P_2(\yhat)$ respectively.
We then perform rescoring to reorder the hypotheses according to the following score:
\begin{equation*}
\log P_{AM}(\yhat | \x) + \alpha_1 \log P_1(\yhat) + \alpha_2 \log P_2(\yhat) + \beta|\yhat|,
\end{equation*}
where $\alpha_1$, $\alpha_2$, $\beta$ are hyper-parameters of the rescoring algorithm optimized on the validation set and $|\yhat|$ is the transcription length in characters (including the spaces between words). In order to diversify the hypotheses in the beam, to increase the probability that the correct transcription is included, we usually relax the threshold in the decoder and increase beam size when dumping beam candidates.

\section{Experiments}
\label{sec:experiments}
\subsection{Technical Details}
\label{sec:experimentstechnicaldetails}

We use the standard splits for \librispeech~and the standard \librispeech~LM corpus for LM training. Models are trained using the wav2letter++\footnote{\url{https://github.com/facebookresearch/wav2letter}} toolkit \cite{pratap2018wav2letter}; reproduction steps and pre-trained models are open-sourced\footnote{\url{https://github.com/facebookresearch/wav2letter}}.

\subsubsection{Acoustic Model Training}
\label{sec:amtraining}
All hyper-parameters including model architecture are cross-validated on \devclean~and \devother. Given that we have a large family of models, for simplicity and clarity, we only report hyper-parameters ranges in which we search their best values.

Plain SGD with momentum is used to train ResNet and TDS models, and Adagrad \cite{duchi2011adaptive} to train Transformers. 
Models are trained on 64 GPUs each with an overall batch size of 256 for ResNet and TDS and 320 for Transformer. 
With only \librispeech, all models converged in under a week; with pseudo-labels from \librivox, training required 2-3 weeks.
The initial learning rate for ResNet models is chosen from [0.05, 0.5]
, while for TDS and Transformer models, the range decreases to [0.01, 0.03]. Specifically, for Transformers, we apply a linear learning rate warm-up schedule for either 32k or 64k updates. For fully-supervised training with \librispeech, the learning rate is halved every 90 epochs for Transformer models, and 150 epochs for ResNet and TDS models. With \librivox, however, we only halve the learning rate once in the middle of the training. For TDS and ResNet models, we use momentum in the range [0.1, 0.6]. With respect to regularization, we use 0.2 dropout everywhere (front-end, encoder, decoder), and layer drop for all Transformer blocks. Dropout in TDS blocks and ResNet convolutions is in the range [0.05, 0.2] and increases with depth. For Seq2Seq training, we run 3 epochs of attention-window pretraining, and use 99\%  of  teacher  forcing (1\%  of uniform output sampling). We also use 10\% dropout in the decoder for TDS (and 0.1 dropout and 0.1 layer drop in the decoder for Transformers), together with 5\% label smoothing, 1\% random sampling and 1\% word piece sampling. 
All models use SpecAugment~\cite{park2019specaug} with an LD policy.

\begin{table}[!t]
\caption{Word-level perplexities of LMs on \librispeech. Perplexity is computed without unknown words.\label{tab:libriPPL}}
\vskip 0.1in
\begin{center}
\begin{small}
\begin{sc}
\begin{tabular}{lcc}
    \toprule
        Language Model & \devclean~ & \devother~ \\
    \midrule
    word 4-gram & 148.0 & 136.6 \\
    ~~~~~~w/o \librivox~& 152.8 & 140.0 \\
    wp 6-gram & 145.4 & 133.7 \\
    wp GCNN (188M) & 61.7 & 61.9 \\
    word GCNN (319M) & 57.0 & 57.9 \\
    word Transf. (562M) & 48.2 & 50.2 \\
    \bottomrule
    \end{tabular}\end{sc}
\end{small}
\end{center}
\vskip -0.1in
\end{table}

\subsubsection{Language Model Training}
\label{sec:lmtraining}
All LMs in this section are trained on the standard \librispeech~LM corpus. All word-level LMs use the same vocabulary for training. $n$-gram LMs are trained with the KenLM toolkit~\cite{heafield2011kenlm}, while the GCNN and Transformer LMs are trained with fairseq toolkit\footnote{\url{https://github.com/pytorch/fairseq}}~\cite{ott2019fairseq}. The word-level 4-gram and GCNN are trained in the same way as \cite{likhomanenko2019needs}. We also train a 6-gram word-piece LM, which has a similar context size to a word-level 4-gram LM, and prunes 5-grams appearing once and 6-gram appearing twice or fewer. The word-piece and word-level GCNN models are trained with Nesterov accelerated gradient descent~\cite{nesterov1983method} on 8 GPUs for 22 epochs with a step-wise learning rate schedule starting from $1$ and decreasing by a factor of 5 when loss stabilized. Gradient clipping and weight normalization are used following~\cite{dauphin2017gcnn}.
The word-level Transformer LM is trained with Nesterov accelerated gradient descent on 128 GPUs for 100 epochs with an inverse square root learning rate schedule. During the first 16000 iterations, a warm-up schedule that linearly increases the learning rate from 0 to 1 is used. Word-level perplexities of all LM variants can be found in Table~\ref{tab:libriPPL}.

\subsection{Results}
\label{sec:results}

\textbf{\librispeech~Results.}~
All our results for \librispeech~are listed in Appendix Table~\ref{tab:libriWERAppendix}. We present results under three scenarios: without any decoding nor external LM, with one-pass decoding only, and with decoding followed by beam rescoring. The decoding beam size is usually 50 and 500 for Seq2Seq and CTC respectively. We use a beam size of 250 for CTC decoding with a GCNN LM.
We train strong baselines on simple ResNet architectures and improve the TDS models significantly compared to past results~\cite{hannun2019TDS}. These convolutional models outperform end-to-end biLSTM models from~\cite{luscher2019transformers}. 
Our best acoustic models are Transformers-based and reach 6.98\% without any decoding on \testother and 5.17\% with decoding and rescoring,
demonstrating that end-to-end training can perform as well as traditional bootstrapped systems.

\begin{table*}[!t]
\caption{WERs on \librispeech~development and test sets. Our best results are shown in the bottom section (with the number of parameters), and are both trained with Seq2Seq loss. Full results can be found in Appendix Table \ref{tab:libriWERAppendix}.\label{tab:libriWER}}
\vskip 0.1in
\begin{small}
    \centering
    \begin{sc}
    \begin{tabular}{lccccccc}
    \toprule
       \multicolumn{2}{c}{AM} & \multicolumn{2}{c}{LM} & \multicolumn{2}{c}{Dev} & \multicolumn{2}{c}{Test} \\
    \cmidrule(lr){1-2} \cmidrule(lr){3-4} \cmidrule(lr){5-6} \cmidrule(lr){7-8}
        \multicolumn{1}{c}{type} & lexicon & type  & lexicon& clean & other & clean & other \\
    \midrule
    Fully Conv. \cite{zeghidour2018fullyconv} & letter & GCNN & word & 3.1 & 9.9 & 3.3 & 10.5 \\
    Conv. Transf. \cite{mohamed2019transformers} & 5k WP & - & - & 4.8 & 12.7 & 4.7 & 12.9 \\
    TDS Conv. \cite{hannun2019TDS} & 10k WP & GCNN & - & 5.0 & 14.5 & 5.4 & 15.6 \\
    ~~~~ Decoding & 10k WP & GCNN & 10k WP & 3.0 & 8.9 & 3.3 & 9.8 \\
    LAS \cite{park2019specaug} & 16k WP & - & - & & & 2.8 & 6.8 \\
    ~~~~ Decoding & 16k WP & RNN & 16k WP & & & 2.5 & 5.8 \\
    biLSTM + attn. \cite{luscher2019transformers} & 10k BPE & - & - & 4.3 & 12.9 & 4.4 & 13.5 \\
    ~~~~ + Transf. decoding & 10k BPE & Transf. & 10k BPE & 2.6 & 8.4 & 2.8 & 9.3 \\
    HMM/biLSTM \cite{luscher2019transformers} & 12k CDp & 4gram+LSTM & word & 2.2 & 5.1 & 2.6 & 5.5 \\
    ~~~~ + Transf. rescoring & 12k CDp & + Transf. & word & 1.9 & 4.5 & 2.3 & 5.0 \\
    Transformers \cite{karita2019comparative} & BPE & RNN & word & 2.2 & 5.6 & 2.6 & 5.7 \\
    Conv. Transf. \cite{han2019stateoftheart} & 6k triphones & 3gram, rescored & word & 1.8 & 5.8 & 2.2 & 5.7 \\
     & & + TDNN + LSTM& & & & & \\
    Conv. Transf. \cite{wang2019transformerbased} & chenones & 4gram & word &  &  & 2.60 & 5.59 \\
    ~~~~ + Transf. rescoring & chenones & Transf. & word & & & 2.26 & 4.85 \\
    \midrule
    Transf. (270M) -- \librispeech & 10k WP  & - & - & 2.54 & 6.67 & 2.89 & 6.98 \\
    ~~~~ + Decoding/Rescoring & 10k WP & GCNN + Transf. & word & 2.10 & 4.79 & 2.33 & 5.17 \\
    Transf. (296M) -- \librivox & 10k WP & - & - & 2.12 & 4.59 & 2.28 & 4.88 \\
    ~~~~ + Decoding/Rescoring & 10k WP  & GCNN + Transf. & word & 2.00 & 3.65 & \textbf{2.09} & \textbf{4.11} \\
    \bottomrule
\end{tabular}
    \end{sc}
\end{small}
\vskip -0.1in
\end{table*}

\textbf{\librivox~Results.}~
Assuming all pseudo-labels are ground-truth, we train acoustic models on a combination of the 960 hours of labeled audio from \librispeech~in conjunction the pseudo-labeled audio from \librivox, where batches are uniformly sampled (without weighting) from both \librispeech~and \librivox~datasets. Transformer AMs with both CTC and Seq2Seq loss were trained for 5 days on this combined dataset, achieving WERs on \testother~of 4.88\% and 2.28\% on \testclean~without decoding or use of an LM, which is state-of-the-art even amongst pipelines that use an LM. Results with decoding/rescoring are shown in Table~\ref{tab:libriWER}, where we reach 2.09\% and 4.11\% on \testclean~and \testother~, respectively, and are further improvements on the state-of-the-art.

\subsection{Ablations}
\label{sec:ablations}

\begin{table}[t]
\caption{WERs of a Transformer AM architecture outlined in section \ref{sec:transformers} trained with Seq2Seq loss on \librispeech~with different amounts of pseudo-labeled audio from \librivox.}
\label{tab:librivoxhoursablation}
\vskip 0.1in
\begin{center}
\begin{small}
\begin{sc}
\begin{tabular}{lcc}
    \toprule
        \shortstack{Training Dataset \\ (Hours)} & \devclean~ & \devother~ \\
    \midrule
    LS \textsc{only} & 2.54 & 6.67 \\
    LS + 1k LV & 2.35 & 5.56 \\
    LS + 3k LV & 2.21 & 5.16 \\
    LS + 10k LV & 2.11 & 4.95 \\
    LS + 53.8k LV & 2.11 & 4.59 \\
    \bottomrule
    \end{tabular}\end{sc}
\end{small}
\end{center}
\vskip -0.1in
\end{table}

\textbf{Varying the amount of unlabeled audio.} ~ In this study, we train on several different randomly-selected subsets of pseudo-labels from the original collection generated as described in Section~\ref{sec:librivoxdataset}. Results are given in Table~\ref{tab:librivoxhoursablation}. Increasing the amount of pseudo-labels strictly improves performance. The listed 53.8k hour result is using the fully-prepared dataset as outlined in Section \ref{sec:librivoxdatasetprep}. WERs given are without decoding after 800k iterations of training.

\textbf{Generating pseudo-labels with an LM containing overlapping text.} ~ As discussed in Section \ref{sec:librivoxdatasetprep}, using an LM to generate pseudo-labels that was trained with a corpus that includes ground truth text from unlabeled audio introduces an overlap that may unrealistically improve the quality of pseudo-labels. We show that the effect of using an LM trained with a small amount of overlapping text to generate pseudo-labels has only a small effect on the performance of models trained on those pseudo-labels.

Table \ref{tab:librivoxlmoverlapablation} contains results for Transformer AMs with both CTC and Seq2Seq loss as described in \ref{sec:transformers} trained on pseudo-labels generated with a decoding step that uses an LM trained on an overlapping versus non-overlapping corpus. The models used are of the same architecture as described in Section \ref{sec:transformers}. There is a small improvement in \devother~ performance for pseudo-labels generated from an overlapping LM, but both models generalize very similarly.

\begin{table}[]
\caption{WERs of a Transformer AM when trained with pseudo-labels generated with a decoder integrating an LM that contains overlapping text with unlabeled audio versus an LM with no overlap. Results are shown after decoding with the word 4-gram language model described in Section
\ref{sec:languagemodels}.\label{tab:librivoxlmoverlapablation}}
\vskip 0.1in
\begin{center}
\begin{small}
\begin{sc}
\begin{tabular}{lccc}
    \toprule
        Model & Overlap & \devother~ & \testother~ \\
    \midrule
    \multirow{2}{*}{Trans. S2S} & No & 4.58 & 4.90 \\
    & Yes & 4.51 & 4.87 \\
    \midrule
    \multirow{2}{*}{Trans. CTC} & No & 4.92 & 5.47 \\
    & Yes & 4.80 & 5.33 \\
    \bottomrule
    \end{tabular}\end{sc}
\end{small}
\end{center}
\vskip -0.1in
\end{table}

\textbf{Training on pseudo-labels only.} ~ Models trained on \librivox~pseudo-labels alone outperform models trained on \librispeech. As outlined in Section~\ref{sec:experiments}, all acoustic models are trained on a combination of \librispeech~and pseudo-labeled \librivox~audio. In this setup, it is difficult to disambiguate the importance of the pseudo-labeled audio compared to supervised data from \librispeech. To test the quality of pseudo-labels in isolation, we trained a CTC-based Transformer model similar to that described in Section \ref{sec:transformers} to compare directly with the CTC-based transformer AM used to generate the pseudo-labels described in Section~\ref{sec:librivoxdataset}. We compare the resulting AM-only performance on the \librispeech~development sets. Without decoding, the resulting \librivox~pseudo-label-only model achieves WERs of 2.38\% and 5.43\% on \devclean~and \devother~respectively, which improves over the \librispeech-only baseline's 2.99\% and 7.31\%, respectively. The volume, quality, and diversity of the generated pseudo-labels alone are sufficient to generate superior results as compared to a model trained only on \librispeech. The model trained on \librispeech~and \librivox~pseudo-labels achieves an improved 2.28\% and 4.99\% on \devclean~and \devother, respectively.

\subsection{End-to-End Acoustic Models Learn a Language Model: Removing the LM from ASR}

In the sections that follow, we show two results. We first give a simple experimental framework to demonstrate that acoustic models trained on speech learn nontrivial language models, and that training on additional audio facilitates learning better acoustic representations. We then show that with a large collection of pseudo-labeled audio, well-trained acoustic models no longer benefit much from decoding with an external language model in most cases.

\textbf{AMs learning LM: transcribing shuffled audio.}~
\label{sec:amlearninglm}
The language modeling properties of end-to-end acoustic models are briefly discussed in \cite{chan2016las}, where an AM trained with CTC is shown to learn an implicit language model based on its predicted posteriors for words with multiple spelling variants. Still other results show that fusing an LM with an AM during training can improve performance \cite{sriram2017cold, chorowski2016towards, wu2016google}. These previous works use RNN-based acoustic models, which possess infinite receptive fields and processes most or all of an input utterance during a single forward pass. We show that modern convolutional architectures have large receptive fields and also condition on internal word representations learned directly from audio.

\begin{figure*}[!b]
\centering
\caption{\label{fig:lm_diffusion} \devother~WERs without decoding across acoustic models and loss functions for original and shuffled versions of \devother~across three settings. Each plot uses the following original and shuffled audio: \textit{Left}: original and shuffled \devother~audio segmented using ASG. \textit{Middle}: audio generated by TTS vocoder for the original and shuffled transcriptions from \devother. \textit{Right}: original and shuffled audio for a subset of \devother~recorded by the paper's authors.}
\begin{tikzpicture}
    \begin{groupplot}[group style = {group size = 3 by 1, horizontal sep = 15pt}, width = 6.4cm, height = 5.cm]
        \nextgroupplot[
            ybar=0.03cm,
            bar width=3pt,
            xtick={1,...,6},
            x tick label style={xshift=2.5ex,rotate=30,anchor=east},
	        ylabel=Word Error Rate,
	        enlargelimits=0.08,
	        ymajorgrids,
 	        xticklabels={ResNet CTC, TDS CTC, Transf. CTC, ResNet S2S, TDS S2S, Transf. S2S}]
            \addplot 
	            coordinates {(1, 10.7517) (2, 12.084) (3, 7.99411) (4, 11.0111) (5, 8.94533) (6, 7.36636)};
            \addplot 
	            coordinates {(1, 6.00519) (2, 6.15252) (3, 5.50556) (4, 5.78099) (5, 5.23973) (6, 5.0956)};
            \addplot+[error bars/.cd, y dir=both, y explicit, error bar style=black]
	            coordinates {
	                (1, 37.62288) +- (0,0.19952) 
	                (2, 41.20872)  +- (0,0.15048)
	                (3, 34.26898)  +- (0,0.32193)
	                (4, 67.21326)  +- (0,3.60783)
	                (5, 37.12714) +- (0,0.23203)
	                (6, 52.18014) +- (0,8.7467)};
            \addplot+[error bars/.cd, y dir=both, y explicit, error bar style=black]
	            coordinates {
	                (1, 29.93562) +- (0,0.2009) 
	                (2, 29.87222)  +- (0,0.21187)
	                (3, 29.51096)  +- (0,0.19126)
	                (4, 29.72806)  +- (0,0.29491)
	                (5, 29.6115) +- (0,0.35259)
	                (6, 30.05988) +- (0,0.30851)};
        \nextgroupplot[
            ybar=0.03cm,
            bar width=3pt,
            x tick label style={xshift=2.5ex,rotate=30,anchor=east},
	        enlargelimits=0.1,
	        ymajorgrids,
	        legend style={at={(0.5,-0.4)},anchor=north,legend columns=-1, legend columns = -1, /tikz/every even column/.append style={column sep=0.5cm}},
 	        xticklabels={ResNet CTC, TDS CTC, Transf. CTC, ResNet S2S, TDS S2S, Transf. S2S},
         	xtick={1,...,6}]
            \addplot 
	            coordinates {(1, 15.5884) (2, 19.1077) (3, 11.0937) (4, 22.6682) (5, 15.2234) (6, 18.5522)};
            \addplot 
	            coordinates {(1, 10.1986) (2, 10.8405) (3, 9.28005) (4, 16.1969) (5, 14.8602) (6, 12.3479)};
            \addplot+[error bars/.cd, y dir=both, y explicit, error bar style=black]
	            coordinates {
	                (1, 36.7159) +- (0,0.04102) 
	                (2, 40.62302)  +- (0,0.16417)
	                (3, 29.19252)  +- (0,0.31493)
	                (4, 41.54312)  +- (0,0.54733)
	                (5, 35.63164) +- (0,0.12032)
	                (6, 37.97166) +- (0,0.36077)};
            \addplot+[error bars/.cd, y dir=both, y explicit, error bar style=black]
	            coordinates {
	                (1, 32.0268) +- (0,0.29151) 
	                (2, 31.6515)  +- (0,0.46834)
	                (3, 27.826)  +- (0,0.57925)
	                (4, 41.04028)  +- (0,0.52823)
	                (5, 39.52852) +- (0,0.645)
	                (6, 32.88842) +- (0,0.29431)};
            \legend{LS original, LS + LV original, LS shuffled, LS + LV shuffled}
        \nextgroupplot[
            ybar=0.03cm,
            bar width=3pt,
            xtick={1,...,6},
            x tick label style={xshift=2.5ex,rotate=30,anchor=east},
	        enlargelimits=0.1,
	        ymajorgrids,
 	        xticklabels={{ResNet CTC}, {TDS CTC}, {Transf. CTC}, {ResNet S2S}, {TDS S2S}, {Transf. S2S}}]
            \addplot 
	            coordinates {(1, 6.20403) (2, 6.5558) (3, 4.70099) (4, 6.97154) (5, 6.5558) (6, 4.50911)};
            \addplot 
	            coordinates {(1, 4.73297) (2, 4.63703) (3, 4.12536) (4, 5.98017) (5, 4.3812) (6, 3.99744)};
            \addplot+[error bars/.cd, y dir=both, y explicit, error bar style=black]
	            coordinates {
	                (1, 16.2136) 
	                (2, 15.1583) 
	                (3, 13.5273) 
	                (4, 14.7106) 
	                (5, 15.9258)
	                (6, 13.6553)};
            \addplot+[error bars/.cd, y dir=both, y explicit, error bar style=black]
	            coordinates {
	                (1, 14.2629) 
	                (2, 13.8152) 
	                (3, 13.2075) 
	                (4, 14.7106) 
	                (5, 14.3588)
	                (6, 12.9837)};
    \end{groupplot}
\end{tikzpicture}
\end{figure*}
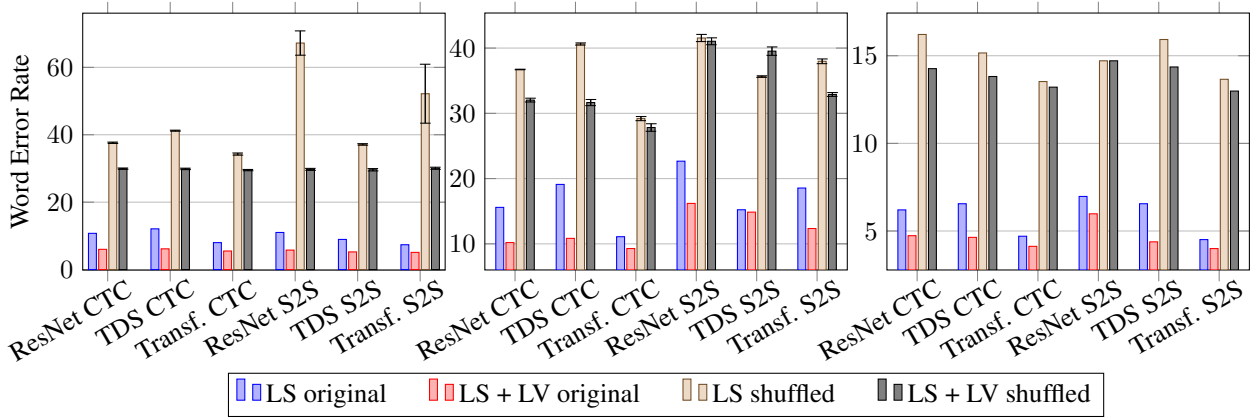

If an AM learns a robust LM, the acoustic model will less effectively predict utterances of high underlying word-perplexity; the model will rely on its acoustic representations to predict words without context, providing a good proxy for the quality of its learned acoustic representations. In the experiments that follow, we introduce a simple ``shuffled transcription'' task in which models transcribe \librispeech~\devother~with utterances corresponding to both unshuffled and shuffled transcriptions. Experiments are performed in three audio settings to eliminate bias when scrambling words. First, with a TTS model, unshuffled and shuffled sentences are forwarded through a WaveRNN vocoder \cite{kalchbrenner2018efficient} trained on the LJSpeech dataset\footnote{\url{https://keithito.com/LJ-Speech-Dataset/}} using the Mozilla TTS toolkit\footnote{\url{https://github.com/mozilla/TTS}}. In the second setting, audio is segmented at the word level using a convolutional stride 2 letter-based AM trained with ASG loss \cite{collobert2016wav2letter}, then re-spliced together in the given shuffled order. Finally, the paper's authors recorded unshuffled and shuffled utterances from a subset of \devother.

Figure~\ref{fig:lm_diffusion} contains the WERs across audio settings on \devother~ without decoding. Both CTC and Seq2Seq models perform poorly across the board on shuffled audio which is expected. As soon as we are interested not in the absolute WER values but in the relative WER values across models / losses / datasets, the main outcome from Figure~\ref{fig:lm_diffusion} is that AMs trained with \librivox~pseudo-labels are able to learn better acoustic representations which improve performance on shuffled inputs for which their internal LMs is not predictive. Full results can be found in Appendix Table \ref{tab:shuffledsentencesappendix}.

\textbf{With enough unlabeled audio, decoding with an LM doesn't improve performance.}~
\label{sec:lmnolm}
The importance of the language model to the success of the pseudo-labeling is known; \cite{kahn2019self} show that in the end-to-end setting, as the quality of the language model used to generate the pseudo-label decreases even marginally, the quality of the model trained on the resulting pseudo-labels decreases. In what follows, we show that through the self-training procedure, decoding an acoustic model trained on \librivox~pseudo-labels generated with the help of a language model gives very small improvements compared to models trained only on \librispeech.

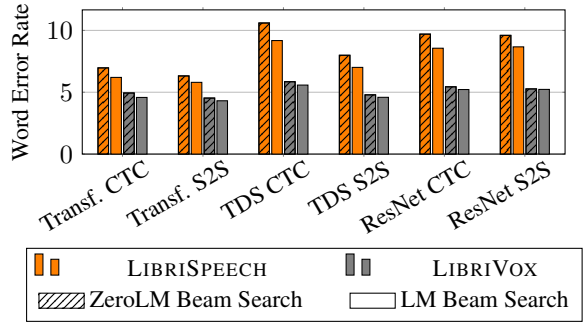
\begin{figure}[!t]
\centering
\caption{\label{fig:lmnolm}WER on \devother~for models trained on \librispeech~and \librispeech~+ \librivox~after decoding with and without the 4-gram LM described in Section \ref{sec:languagemodels}. The gain from LM beam-search decoding for models trained on \librivox~ is much smaller compared to that for models trained on \librispeech.}
\begin{tikzpicture}[]
    \begin{axis}[
        ybar=0.03cm,
        width=8cm,
        height=3.5cm,
        bar width=4pt,
        ymajorgrids,
        ylabel={Word Error Rate},
        ymin=0,
        xtick=data,
        x tick label style={xshift=2.5ex,rotate=30,anchor=east,yshift=-2pt,font=\small},
        y label style={font=\small},
        symbolic x coords = {Transf. CTC, Transf. S2S, TDS CTC, TDS S2S, ResNet CTC, ResNet S2S},
        tickwidth         = 1pt,
        legend style={
            at={(0.46,-0.65)},
            anchor=north,
            legend columns=2,
            font=\small,
            /tikz/every even column/.append style={column sep=0.5cm},
        },
        ]
        
        \addplot[black, fill=orange, postaction={pattern=north east lines}] coordinates { 
            (Transf. CTC, 6.97)
            (Transf. S2S, 6.32)
            (TDS CTC, 10.60)
            (TDS S2S, 7.99)
            (ResNet CTC, 9.70)
            (ResNet S2S, 9.60)
        };

        \addplot[black, fill=orange] coordinates {
            (Transf. CTC, 6.20)
            (Transf. S2S, 5.80)
            (TDS CTC, 9.18)
            (TDS S2S, 7.01)
            (ResNet CTC, 8.56)
            (ResNet S2S, 8.67)
        };

        \addplot[black, fill=gray, postaction={pattern=north east lines},
            area legend,
            legend image post style={draw opacity=0, color=white}
        ] coordinates {
            (Transf. CTC, 4.94)
            (Transf. S2S, 4.53)
            (TDS CTC, 5.85)
            (TDS S2S, 4.79)
            (ResNet CTC, 5.44)
            (ResNet S2S, 5.27)
        };
        
        \addplot[black, fill=gray] coordinates {
            (Transf. CTC, 4.58)
            (Transf. S2S, 4.31)
            (TDS CTC, 5.58)
            (TDS S2S, 4.59)
            (ResNet CTC, 5.22)
            (ResNet S2S, 5.23)
        };
                    
        \legend{,,,}
        \addlegendimage{black, fill=orange}
        \addlegendentry{\librispeech}
        \addlegendimage{black, fill=gray}
        \addlegendentry{\librivox}
        \addlegendimage{area legend,pattern=north east lines}
        \addlegendentry{ZeroLM Beam Search}
        \addlegendimage{area legend, pattern=north east lines, color=black, fill=white}
        \addlegendentry{LM Beam Search}
    \end{axis}
\end{tikzpicture}
\end{figure}

Results are shown in Figure~\ref{fig:lmnolm}. We use a beam-search decoding procedure without an LM (``Zero-LM'') to disambiguate the effect of beam search on WER, and evaluate on \devother~ to provide a better lower bound for how much decoding with the LM can improve performance (decoder parameters are also optimized on \devother~). The models for which results are shown are trained on pseudo-labels from \librivox~generated with an $n$-gram language model without an overlapping text corpus (see the ablation in Sections \ref{sec:ablations} and \ref{sec:languagemodels}). Decoding with the LM gives little to no gain for models trained on \librispeech~+ \librivox~and a much more significant gain for those models trained only on \librispeech, suggesting information from the 4-gram LM used to generate pseudo-labels on \librivox~has thoroughly diffused into AMs trained with those labels. Full results can be found Appendix in Table \ref{tab:libriWERAppendix}.

\section{Related Work} 
Deep neural networks were reintroduced in ASR with HMMs \cite{hinton2012dnnasr}, and many of state-of-the-art models still rely on force alignment \cite{han2017capio,luscher2019transformers,karita2019comparative}. Nonetheless, there have been increasingly competitive end-to-end results trained with CTC \cite{graves2014e2easr,amodei2016deep}, ASG \cite{collobert2016wav2letter, zeghidour2018fullyconv}, LF-MMI \cite{hadian2018end}, sequence-to-sequence \cite{chan2016las,chiu2018s2s}, transduction \cite{prabhavalkar2017comparison,he2019rnnt}, and differentiable decoding \cite{collobert2019diffdec}. \textit{Listen Attend and Spell} \cite{chan2016las} is a family of end-to-end models based on biLSTMs which achieved state-of-the-art results with improved regularization through data augmentation \cite{park2019specaug}; we consequently use SpecAugment in all of our experiments. Seq2Seq models are not limited to RNNs; time-depth separable convolutions also give strong results \cite{hannun2019TDS}. Our best models are transformer-based, as \cite{luscher2019transformers,karita2019comparative}, which give good results in Seq2Seq settings even without external LMs \cite{mohamed2019transformers}. In ASR, semi-supervised pseudo-label-style self-training has been explored generally in end-to-end settings in~\cite{soltau2016neural,li2019semi,kahn2019self} for both low-resource~\cite{vesely2017semi,cui2017knowledge} and large-scale~\cite{parthasarathi2019lessons} setups.

\section{Discussion}
\label{sec:discussion}
We presented state-of-the-art results on \librispeech~with end-to-end methods.
While allowing for lexicon-free decoding, the 10k word-piece tokens used during training limit the amount of striding we can use in our model architectures and can be replaced by AMs outputting words with an arbitrary lexicon \cite{collobert2019word}. As relative WER gains due to language models shrink (from $\approx$20\% relative-WER without \librivox~to $\approx$10\% with, for GCNN decoding), and as we showed that AMs learn LM-level information, differentiable decoding \cite{collobert2019diffdec}
is a possible avenue for single-stage AM + LM joint training.

We show the effectiveness of a simple pipeline
that does not require many training steps. In light of our semi-supervised results without decoding or an LM, we think Seq2Seq/CTC losses, transducers, and differentiable decoding are viable methods to achieve end-to-end state-of-the-art results, without external LMs, through semi-supervised learning.

\section{Acknowledgements}
We would like to thank Steven Garan for audio recordings of shuffled sentences from \librispeech{} \devother{}.

\appendix

\begin{table*}[!t]
\caption{Word error rates on \librispeech's development and test sets. Our models listed in the middle and bottom blocks are trained with CTC and Seq2seq losses respectively.\label{tab:libriWERAppendix}}
\begin{small}
    \centering
    \begin{sc}
    \begin{tabular}{lccccccc}
    \toprule
       \multicolumn{2}{c}{AM} & \multicolumn{2}{c}{LM} & \multicolumn{2}{c}{Dev} & \multicolumn{2}{c}{Test} \\
    \cmidrule(lr){1-2} \cmidrule(lr){3-4} \cmidrule(lr){5-6} \cmidrule(lr){7-8}
        \multicolumn{1}{c}{type} & lexicon & type  & lexicon& clean & other & clean & other \\
    \midrule
      \multicolumn{1}{c}{ \textbf{CTC} } & \\
    ResNet (306M) & 10k WP & - & - & 3.93 & 10.13 & 4.08 & 10.03 \\
    ~~ Decoding & & ZeroLM & lex & 3.76 & 9.7 & 4.07 & 9.77 \\
    ~~ Decoding & & 4gram & word & 3.29 & 8.56 & 3.68 & 8.69 \\
    ~~ Decoding & & GCNN & word & 2.99 & 7.50 & 3.28 & 7.53 \\
    ResNet (500M) \librivox~& 10k WP & - & - & 2.34 & 5.54 & 2.55 & 5.99 \\
    ~~ Decoding & & ZeroLM & lex & 2.37 & 5.45 & 2.73 & 5.96 \\
    ~~ Decoding & & 4gram & word & 2.34 & 5.23 & 2.68 & 5.75 \\
    ~~ Decoding & & GCNN & word & 2.19 & 4.64 & 2.45 & 5.13 \\
    \cmidrule(lr){1-1}
    TDS (200M) & 10k WP & - & - & 4.22 & 11.16 & 4.63 & 11.16 \\
    ~~ Decoding & & ZeroLM & lex & 3.93 & 10.61 & 4.44 & 10.67 \\
    ~~ Decoding & & 4gram & word & 3.49 & 9.18 & 3.98 & 9.53 \\
    ~~ Decoding & & GCNN & word & 2.92 & 7.52 & 3.40 & 8.05 \\
    TDS (500M) \librivox~& 10k WP & - & - & 2.44 & 5.70 & 2.66 & 6.11 \\
    ~~ Decoding & & ZeroLM & lex & 2.47 & 5.61 & 2.86 & 6.18 \\
    ~~ Decoding & & 4gram & word & 2.44 & 5.33 & 2.81 & 5.91 \\
    ~~ Decoding & & GCNN & word & 2.26 & 4.71 & 2.55 & 5.24 \\
    \cmidrule(lr){1-1}
    Transf. (322M) & 10k WP & - & - & 2.99 & 7.31 & 3.09 & 7.40 \\
    ~~ Decoding & & ZeroLM & lex & 2.85 & 6.98 & 3.14 & 7.23 \\
    ~~ Decoding & & 4gram & word & 2.63 & 6.20 & 2.86 & 6.72 \\
    ~~~~ + Rescoring &  & GCNN + Transf. & word & 2.18 & 4.90 & 2.44 & 5.36 \\
    ~~ Decoding & & GCNN & word & 2.35 & 5.29 & 2.57 & 5.85 \\
    ~~~~ + Rescoring &  & GCNN + Transf. & word & 2.20 & 4.94 & 2.47 & 5.45 \\
    Transf. (299M) \librivox~& 10k WP & - & - & 2.28 & 5.00 & 2.39 & 5.35  \\
    ~~ Decoding & & ZeroLM & lex & 2.31 & 4.94 & 2.58 & 5.42 \\
    ~~ Decoding & & 4gram & word & 2.24 & 4.59 & 2.52 & 5.22 \\
    ~~~~ + Rescoring &  & GCNN + Transf. & word & 1.99 & 3.91 & 2.28 & 4.50 \\
    ~~ Decoding & & GCNN & word & 2.09 & 4.27 & 2.41 & 4.79 \\
    ~~~~ + Rescoring & & GCNN + Transf. & word & 2.01 & 3.95 & 2.31 & 4.54 \\
    \midrule
      \multicolumn{1}{c}{ \textbf{Seq2Seq} } & \\
    ResNet (389M) & 10k WP & - & - & 3.51 & 9.89 & 4.92 & 10.33 \\
    ~~ Decoding & & ZeroLM & lexfree & 3.42 & 9.60 & 4.31 & 9.59 \\
    ~~ Decoding & & 6gram & 10k WP & 3.05 & 8.69 & 3.88 & 8.88 \\
    ~~ Decoding & & GCNN & 10k WP & 2.78 & 7.86 & 3.79 & 8.21 \\
    ResNet (500M) \librivox~& 10k WP & - & - & 2.27 & 5.29 & 2.86 & 5.88 \\
    ~~ Decoding & & ZeroLM & lexfree & 2.26 & 5.28 & 2.67 & 5.54 \\
    ~~ Decoding & & 6gram & 10k WP & 2.29 & 5.25 & 2.69 & 5.62 \\
    ~~ Decoding & & GCNN & 10k WP & 2.26 & 4.91 & 2.66 & 5.31 \\
    \cmidrule(lr){1-1}
    TDS (190M) & 10k WP & - & - & 3.20 & 8.20 & 3.43 & 8.30 \\
    ~~ Decoding & & ZeroLM & lexfree & 2.89 & 8.00 & 3.24 & 7.99 \\
    ~~ Decoding & & 6gram & 10k WP & 2.76 & 7.01 & 3.18 & 7.16 \\
    ~~ Decoding & & GCNN & 10k WP & 2.54 & 6.30 & 2.93 & 6.43 \\
    TDS (500M) \librivox~& 10k WP & - & - & 2.17 & 4.78 & 2.37 & 5.15 \\
    ~~ Decoding & & ZeroLM & lexfree & 2.20 & 4.80 & 2.38 & 5.11 \\
    ~~ Decoding & & 6gram & 10k WP & 2.18 & 4.61 & 2.35 & 5.02 \\
    ~~ Decoding & & GCNN & 10k WP & 2.08 & 4.21 & 2.24 & 4.61 \\
    \cmidrule(lr){1-1}
    Transf. (270M) & 10k WP & - & - & 2.54 & 6.67 & 2.89 & 6.98 \\
    ~~ Decoding & & ZeroLM & lexfree & 2.49 & 6.32 & 2.75 & 6.58 \\
    ~~ Decoding & & 6gram & 10k WP & 2.29 & 5.81 & 2.72 & 6.23 \\
    ~~~~ + Rescoring &  & GCNN + Transf. & word & 2.13 & 5.00 & 2.51 & 5.47 \\
    ~~ Decoding & & GCNN & 10k WP & 2.12 & 5.20 & 2.40 & 5.70 \\
    ~~~~ + Rescoring &  & GCNN + Transf. & word & 2.10 & 4.79 & 2.33 & 5.17 \\
    Transf. (296M) \librivox~& 10k WP & - & - & 2.12 & 4.59 & 2.28 & 4.88 \\
    ~~ Decoding & & ZeroLM & lexfree & 2.10 & 4.53 & 2.27 & 4.80  \\
    ~~ Decoding & & 6gram & 10k WP & 2.06 & 4.32 & 2.25 & 4.70 \\
    ~~~~ + Rescoring &  & GCNN + Transf. & word & 1.91 & 3.76 & 2.10 & 4.20 \\
    ~~ Decoding & & GCNN & 10k WP & 1.97 & 3.95 & 2.17 & 4.37 \\
    ~~~~ + Rescoring &  & GCNN + Transf. & word & 2.00 & 3.65 & 2.09 & 4.11 \\
    \bottomrule
    \end{tabular}
    \end{sc}
\end{small}
\end{table*}

\clearpage

\section{Experiment Details}
Comprehensive WER results for \librispeech{} and \librivox{} acoustic models, including with greedy and beam-search decoding with different LMs and beam rescoring can be found in Table~\ref{tab:libriWERAppendix}. This section mainly focus on providing details of how we optimize the beam-search decoding and rescoring procedures for our acoustic models. 

\subsection{Beam-Search Decoding}
When beam-search decoding, we use the \devclean{} and \devother{} sets as validation sets and use random search to optimize decoding hyper-parameters. The search ranges of those hyper-parameters are listed in Table~\ref{tab:bsdparamsappendix}. We use between 64 and 128 runs in each random search with hyper-parameter values uniformly sampled from the given ranges. It is worth noting that the optimal ranges for language model weight for models trained on \librispeech{} are higher than ones found for \librivox{} models as shown in Table~\ref{tab:bsdoptimalappendix}. This is conceivably additional evidence that models trained with additional audio rely less on language models.

\begin{table}[h!]
\caption{Hyper-parameter values and ranges used in a random search for beam-search decoding with $n$-gram (top block) and GCNN (bottom block) LMs.\label{tab:bsdparamsappendix} }
\vskip 0.1in
\begin{center}
\begin{small}
\begin{sc}
\begin{tabular}{lcccc}
    \toprule
    & \multicolumn{2}{c}{\librispeech{}} & \multicolumn{2}{c}{\librivox{}} \\
    \multicolumn{1}{c}{Parameters} & CTC & S2S & CTC & S2S \\
    \midrule
    beam & $500$ & $50,100$ & $500$ & $20,50,100$\\
    token beam & $100$ & $10,50$ & $100$ & $3,5,10$ \\
    LM weight & $[0, 3]$ & $[0, 2]$ & $[0, 1.5]$ & $[0, 1]$ \\
    threshold & $100$ & $10,50$ & $100$ & $5,10,50$ \\
    word insert. & $[-3, 3]$ & - & $[-3, 3]$ & - \\
    EOS-penalty & - & $[-10, 0]$ & - & $[-10, 0]$ \\
    \midrule
    beam & $250$ & $50$ & $250$ & $20,50,100$\\
    token beam & $100$ & $10,18$ & $100$ & $3,5,10$ \\
    LM weight & $[0, 3]$ & $[0, 2]$ & $[0, 1.5]$ & $[0, 0.8]$\\
    threshold & $20$ & $10,15$ & $20$ &  $5,10,50$ \\
    word insert. & $[-3, 3]$ & - & $[-3, 3]$ & - \\
    EOS-penalty & - & $[-10, 0]$ & - & $[-10, 0]$ \\
    \bottomrule
    \end{tabular}\end{sc}
\end{small}
\end{center}
\vskip -0.1in
\end{table}

\begin{table}[h!]
\caption{Optimal LM weight ranges (based on WER) for beam-search decoding with $n$-gram (top block) and GCNN (bottom block) LMs found via random search. \label{tab:bsdoptimalappendix} }
\vskip 0.1in
\begin{center}
\begin{small}
\begin{sc}
\begin{tabular}{lcccc}
    \toprule
    & \multicolumn{2}{c}{\librispeech{}} & \multicolumn{2}{c}{\librivox{}} \\
    \multicolumn{1}{c}{Data} & CTC & S2S & CTC & S2S \\
    \midrule
    clean & $[0.8, 1.4]$ & $[0.6, 1.1]$ & $[0.2, 0.4]$ &  $[0.0, 0.2]$ \\
    other & $[1.1, 1.9]$ & $[0.6, 1.2]$ & $[0.5, 0.7]$ & $[0.1, 0.5]$ \\
    \midrule
    clean & $[0.4, 0.8]$ & $[0.2, 0.5]$ & $[0.2, 0.5]$ & $[0.0, 0.4]$ \\
    other & $[0.5, 1.1]$ & $[0.3, 0.7]$ & $[0.3, 0.6]$ & $[0.2, 0.4]$ \\
    \bottomrule
    \end{tabular}\end{sc}
\end{small}
\end{center}
\vskip -0.1in
\end{table}

\subsection{Rescoring}
To perform rescoring, we first dump all hypotheses proposed during beam-search decoding using the optimal hyper-parameters found with random search. When dumping candidates, beam size, token beam size, and beam threshold are increased so as to increase the number of proposed hypotheses on which to run rescoring. Further details are listed in Table~\ref{tab:rescoreappendix}. We find optimal values of rescoring hyper-parameters $\alpha_1$, $\alpha_2$ and $\beta$ (see Section \ref{sec:rescoring}) via a grid search for CTC models ($\alpha_1,\beta\in[0,1]$ and $\alpha_2\in[-0.3, 0.3]$ where the grid step is set to $0.1$), and a random search for sequence-to-sequence models ($\alpha_1,\in[0,2.5]$, $\alpha_2\in[-1, 1]$, $\beta\in[-3,3]$ with 1000 attempts).

\begin{table}[h!]
\caption{Parameters values used when dumping beam candidates for rescoring with $n$-gram (top block) and GCNN (bottom block) LMs.\label{tab:rescoreappendix} }
\vskip 0.1in
\begin{center}
\begin{small}
\begin{sc}
\begin{tabular}{lcc}
    \toprule
    Parameters & CTC & S2S \\
    \midrule
    beam & $2500$ & $250$ \\
    token beam & $1500$ & $150$ \\
    threshold & $5000$ & $150$ \\
    \midrule
    beam & $250$ & $250$ \\
    token beam & $100$ & $100$ \\
    threshold & $20$ & $100$ \\
    \bottomrule
    \end{tabular}\end{sc}
\end{small}
\end{center}
\vskip -0.1in
\end{table}

\section{Generating Shuffled Audio}
This section provides details of how we generated shuffled utterances used in the experiments in Section~\ref{sec:amlearninglm}. Each experiment could introduce systematic error. Therefore, we propose several experiments to conclude. For the two methods generating existing or using new audio (\textbf{TTS} and \textbf{Segmentation}), we shuffle \devother{} five times and report the mean and standard deviation (as error bars) in Figure~\ref{fig:lm_diffusion}.

\subsection{TTS}
For each sentence in \devother{}, we randomly shuffle its words to form a new sentence. We run the resulting text through a TTS model as outlined in Section~\ref{sec:ablations} to create synthetic audio for the scrambled sentences. While simple and easy to implement, this method introduces and amplifies intrinsic errors in the TTS model into the ablation. In particular, the model struggles to handle many of the rare words present in \devother{}. Also TTS approach is still away from the human speech.

\subsection{Segmentation}
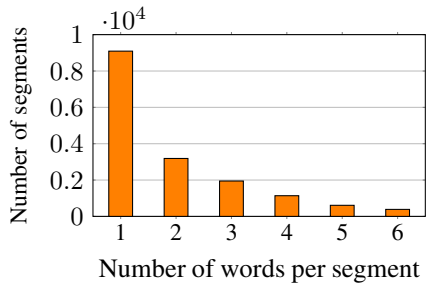
\begin{figure}[h!]
\centering
\begin{tikzpicture}[]
    \begin{axis}[
        ybar=0.03cm,
        width=6cm,
        height=4cm,
        bar width=9pt,
        ymajorgrids,
        ylabel={Number of segments},
        xlabel={Number of words per segment},
        ymin=0,
        xtick=data,
        x tick label style={xshift=1.3ex,rotate=0,anchor=east,yshift=-5pt,font=\small},
        y label style={font=\small},
        symbolic x coords = {1, 2, 3, 4, 5, 6},
        tickwidth         = 1pt,
        ]
        \addplot[black, fill=orange] coordinates {
            (1, 9094)
            (2, 3188)
            (3, 1947)
            (4, 1138)
            (5, 610)
            (6, 385)
        };
    \end{axis}
\end{tikzpicture}
\caption{\label{fig:segngramappendix}Distribution of all $n$-grams in the obtained segments of filtered \devother{} (1969 samples with 16,362 segments in total).}
\end{figure}

With this method, we first force-align the transcriptions of \devother{} to the existing audio using a letter-based stride-two ASG model as outlined in Section~\ref{sec:ablations} and collecting the beginning timestamp and duration of each word. Then, to avoid splicing words that are ready closely together, audio samples are only split when silence of longer than 130 milliseconds is detected (split is done in the middle of silence segment). Finally, audio chunks are randomly shuffled and re-assembled into new utterances. Since this ablation aims to remove LM-friendly context from audio, we filter the resulting recombined audio samples. In particular, we filter all utterances that have only one segment, or have at least one segment with more than 6 words in it. After filtering, 1969 out of 2864 samples in \devother{} remain. The distribution of the number of words in each of the resulting segments is shown in Figure~\ref{fig:segngramappendix}. 

Unlike the TTS method described above, the segmentation method reuses audio as much as possible from \devother{}. That said, neither the force alignment nor the segmentation techniques handle all the word boundaries. As such, there may be incomplete words in the resulting audio and LM-friendly context.    

\subsection{Recording}
The paper's authors recorded 184 randomly selected sentences from \devother{} as well as a single set of shuffled utterances. The unshuffled recorded audio has the lowest WER among all the three methods. We plan to complete a collection of unshuffled and shuffled audio for \devother{} in future work.

\begin{table}[h!]
\caption{Performance of word-level 4-gram and Transformer LMs from Table \ref{tab:libriPPL} on original and shuffled audio transcriptions generated from \librispeech{} \devother{}.}
\label{tab:shuffledsentencesappendix}
\vskip 0.1in
\begin{center}
\begin{small}
\begin{sc}
\begin{tabular}{lccc}
    \toprule
    Setting & Shuffled & 4-gram LM & Transf. LM \\
    \midrule
    TTS & No & 147 & 50 \\
    TTS & Yes & 749 $\pm$ 2 & 389 $\pm$ 2 \\
    \midrule
    Segment. & No & 167 & 56 \\
    Segment. & Yes & 827 $\pm$ 5 & 743 $\pm$ 9 \\
    \midrule
    Recording & No & 162 & 49 \\
    Recording & Yes & 3807 & 2995 \\
    \bottomrule
    \end{tabular}\end{sc}
\end{small}
\end{center}
\vskip -0.1in
\end{table}

\subsection{Perplexity}
As shown in Table~\ref{tab:shuffledsentencesappendix}, there are large gaps between the perplexity of transcriptions in the original and shuffled sets across all settings. Our shuffling strategy thus removes important word context and breaks the alignment of the audio words distribution with the LM. The WER gap between the two sets is thus a proxy for the amount of language modeling an acoustic model may implicitly perform.

\end{document}